\documentclass[letterpaper]{article}
\usepackage{iccc}

\usepackage{times}
\usepackage{helvet}
\usepackage{url}
\usepackage{courier}
\usepackage{latexsym}
\usepackage{graphicx}
\usepackage{multirow}
\usepackage{xspace,mfirstuc,tabulary}
\usepackage{vcell}
\usepackage{colortbl}
\title{Creative Data Generation: A Review Focusing on Text and Poetry}
\author{Mohamad Elzohbi, Richard Zhao \\
Department of Computer Science, University of Calgary \\ Calgary, Alberta, Canada T2N 1N4  \\
\texttt{\{melzohbi, richard.zhao1\}@ucalgary.ca}}

\begin{document} 
\maketitle
\begin{abstract}
\begin{quote}
The rapid advancement in machine learning has led to a surge in automatic data generation, making it increasingly challenging to differentiate between naturally or human-generated data and machine-generated data. Despite these advancements, the generation of creative data remains a challenge. This paper aims to investigate and comprehend the essence of creativity, both in general and within the context of natural language generation. We review various approaches to creative writing devices and tasks, with a specific focus on the generation of poetry. We aim to shed light on the challenges and opportunities in the field of creative data generation.
\end{quote}
\end{abstract}

\section{Introduction} 
Data refers to information that can be stored and processed by a computer. It can take many forms and can be generated both naturally, such as through the thoughts and ideas in a person's mind, and artificially, such as through the use of machine-based models. In particular, computer scientists use the term data to refer to anything that can be stored in the computer's memory called binary data. The process of generating data can be (1) fully automated, like using a generative model to generate poetry from topic words, (2) semi-automated, where the output is a collaboration between the machine and the human, such as a poem draft generated with human refinements \cite{lamb2017taxonomy}, or (3) entirely manual such as using a text editor to write a poem.

The concept of creativity has historically been difficult to define and has not been considered seriously in AI, as it was thought that machines need first to be capable of possessing thoughts and experiencing emotions \cite{colton2012computational}. With the advent of deep learning and transformer models, natural language generation (NLG) techniques have become more advanced and the possibility of generating creative output seems more viable.

This paper focuses on the intersection between creative data generation and NLG with an emphasis on poetry generation.
First, we address the difficulties in defining creativity and present our own perspective on the key elements of creativity. Second, we review relevant metrics used and how they are relevant to the proposed criteria, then we evaluate NLG models based on those criteria. Third, we provide an overview of the practical creative applications of text generation tasks, reviewing some of the most recent work in this area. We focus on the poetry generation task, examining the methods and models employed.

\section{Creativity Paradox}
\label{section:2}
Creativity is a complex and often-debated concept that is far from being well-defined. While many tend to focus on the creativity of the end result, the process by which it is created may also play a role in determining its level of creativity. When the steps involved in generating an output are clearly defined and easy to replicate, the output may be seen as less creative, even if it was initially considered as such. 
For example, the creativity of Johannes Vermeer, one of the most well-known Dutch artists in western history, was put onto the table of controversial discussions when Philip Steadman suggested that Vermeer was using optics in the painting process \cite{steadman2002:vermeer,timvermeer}.
Furthermore, the English painter \citeauthor{hockney_secret_2006} (\citeyear{hockney_secret_2006}) argued that most painters since the 15th century secretly used their knowledge of optical science in the painting process.
It could be questioned as to why certain painters would choose to keep their techniques secret if this does not raise doubts about the authenticity of their artistic talent and creativity. One possible explanation is that artists desire for their work to be perceived as unique and distinguished from craft-based forms of expression. They may view their work as constituting a higher form of artistic expression, and therefore seek recognition as elite artists rather than as craftsmen, who are considered to have a lower social status \cite{markowitz1994distinction}. This indicates that the level of creativity in the output can be significantly impacted by the degree of unpredictability. Hence, can photography be considered creative? Paul Delaroche, a classical painter, believed that the emergence of photography marked the death of painting. However, painters disagree and see painting as expressive and original, as it is not simply a mirror reflection of the real world. \cite{crimp_end_1981}. 
On the other side, pictorialists argue that they can express their vision through photography by adding their own touches to a real photograph \cite{hertzmann_can_2018}. Consequently, how valid is it to attribute creativity to machines if they are simply following a predetermined set of rules written by humans? In fact, even when we have a clear understanding of the design of a machine, we may not fully comprehend the internal processes that occur during its training. Does the hint of anonymity allow for the possibility of attributing creativity to them? Does revealing the process diminish the perceived originality of its output? Do we, as humans, understand creativity based on beautifulness or wonderfulness of the output? 

\citeauthor{shneiderman2007creativity} (\citeyear{shneiderman2007creativity}) summarizes the creative process into three main schools: (1) structuralists: who outline the creative process in four stages: (i) gathering information (preparation), (ii) forming new connections (incubation), (iii) finding sudden insights (illumination) and (iv) verifying and refining those insights (verification). (2) inspirationalists: who believe that creative insights can be achieved through sketching, visualization and meditation. (3) situationalists: who see creativity as a social process. In \citeauthor{boden2004creative} (\citeyear{boden2004creative})'s point of view, creativity must induce (i) new, (ii) surprising, and (iii) valuable ideas or artifacts. As a situationalist, \citeauthor{csikszentmihalyi1997flow} (\citeyear{csikszentmihalyi1997flow}) asserts that novelty is not enough to be considered creative, but the work must be accepted by the relevant field. 
\citeauthor{hertzmann_can_2018} (\citeyear{hertzmann_can_2018}) argues that machines are merely tools in the hands of artists. These tools are not always predictable but not inherently creative. \citeauthor{hertzmann_can_2018} argues that just as we do not attribute creativity to a brush when watercolors flow on a canvas or to nature and animals when they form beautiful patterns and stunning structures, although animals have brains just like humans, we should not attribute it to machines. 

\begin{figure}
	\centering
	\includegraphics[width=0.3\textwidth]{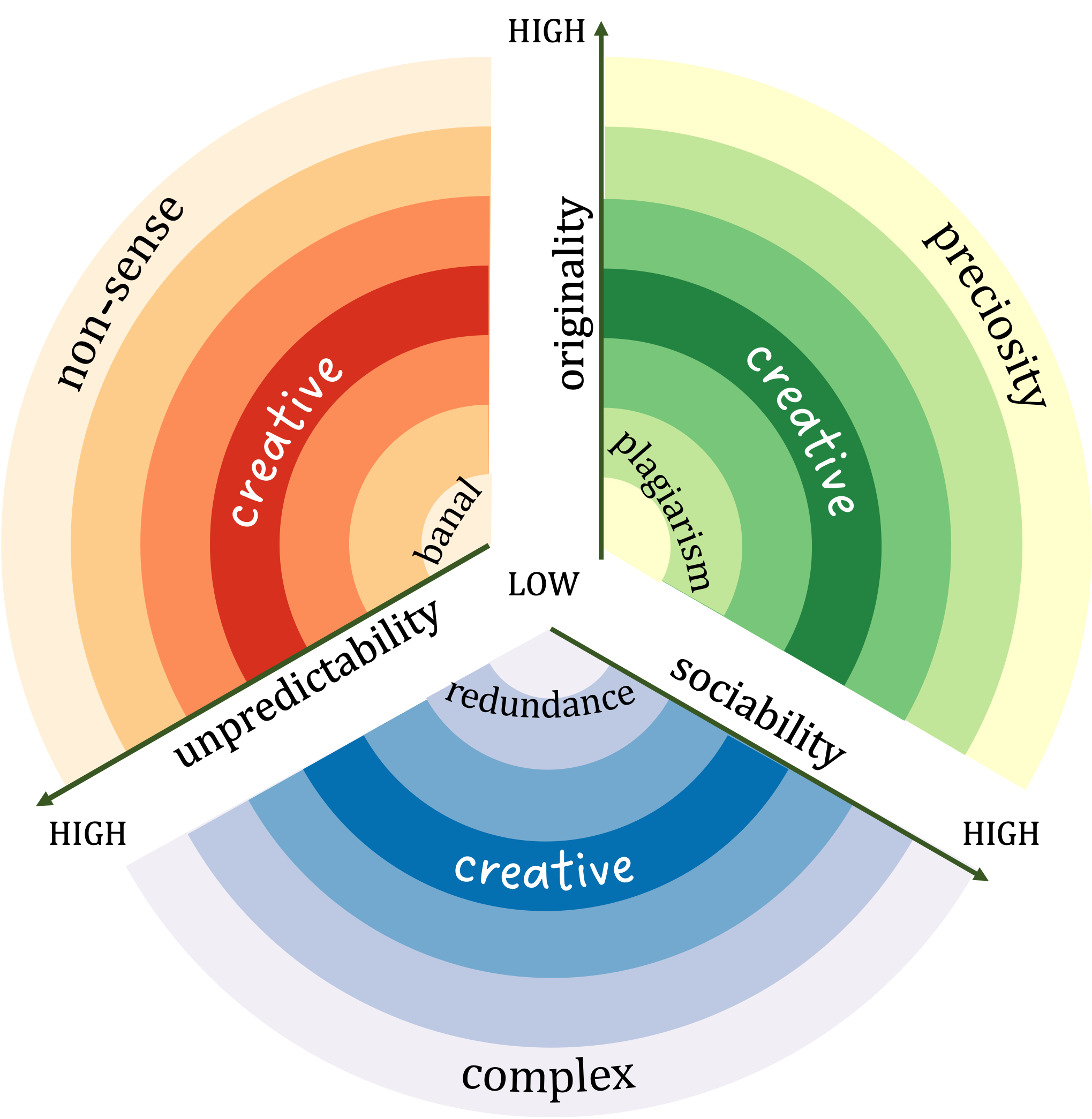}
	\caption{Creativity Dimensions. \nocite{lamb2017taxonomy}}
	\label{fig:creativity}
  	\vspace{-9pt}
\end{figure}

Ultimately, the concept of creativity appears to be complex and multifaceted, making it seem impossible to develop a cohesive paradigm. However, we can see recurring themes and intersecting dimensions that we will summarize in the following: (see Figure \ref{fig:creativity}) 

\begin{enumerate}
    \item \textbf{Originality}: To be considered creative, an output must strike a balance in originality. To achieve this, a creative model must avoid lacking originality (plagiarism) especially when learning from creative works. The greater the originality in the output, the more creative it is considered to be. However, too much originality results in confusion, disconnection, and excessive refinement (preciosity).

    \item \textbf{Unpredictability}: An output that is obvious and predictable is generally seen as banal and boring. The greater the surprise factor in the output, the more creative it is considered to be. However, the output that is completely unpredictable may be perceived as nonsensical and lose its creativity. To achieve a balance between surprise and coherence, a creative model may need to learn a human-like probability distribution and optimize its decoding strategies for creative expression.
    
    \item \textbf{Sociability}: We define sociability as the density of meaning and imagery in the output in relation to its appropriate domain. Poets and painters, for example, use language and colors, respectively, to compact complex ideas and emotions into a limited space, such as a verse or a canvas. A work that is semantically condensed provides more material for the artist or critic to engage with and assess. If the work is less sociable, it may be seen as redundant. However, it is important for the output to strike a balance, as an output that is overly dense may be complex and difficult to comprehend.
\end{enumerate}

\section{Creative NLG: Aspects}
\label{section:3}
Discussions about NLG in the past have centered around determining what constitutes as \textit{real} NLG \cite{areiter2016}. As the field progresses, the focus has shifted towards determining what constitutes as \textit{creative} NLG. 
We believe that a creative NLG model must exhibit a sufficient level of originality, sociability, and unpredictability.

Despite the significant advances in NLG, the study of the originality in language models has received relatively little attention. Originality in language can take various forms, such as phrasal, thematic, semantic, or stylistic originality, among others. Although probabilistic models may lead to the generation of original phrases, there is also the risk of over-fitting and copying from the training corpus. Plagiarism in creative language is particularly noticeable, as the output deviates from the normal use of language. This has been, recently, a concern in other creative fields, such as in music generation \cite{yin2021good}. \citeauthor{brooks2021got} (\citeyear{brooks2021got}) have noted that there is currently no standard automatic test for originality in NLG and proposed an approach to test the original use of language to identify plagiarism violations using a phrasal counting approach in the ground truth. However, this approach is limited to measuring the phrasal originality which is also important to identify phrases that diverted from their connotation and lost their imagery due to repetitive use in ordinary speech, such as dead metaphors.

\citeauthor{shannon1951prediction} (\citeyear{shannon1951prediction}) conducted experiments to predict language redundancy by asking humans to guess the next character in an English passage.  \citeauthor{shannon1951prediction} assumed that humans sample from a probability distribution based on their previously learned letter-sequence probabilities and introduced the concept of language entropy. \[H = - \sum\nolimits_{{i}}^{}  {p\left( {{w_i}} \right)\ \log \left( {p\left( {{w_i}} \right)} \right)}\] where $p(w_i)$ is the probability for the $i^{th}$ n-gram to occur. \citeauthor{paisley1966effects} (\citeyear{paisley1966effects}) studied the effect of several factors, including structure, on the redundancy of the language and found that prose is more redundant than poetic language. A more redundant language is less dense hence less sociable. Due to form restrictions in poems, it seems at first glance that the population of the allowed text must decrease from that of prose, but \citeauthor{manin2021running} (\citeyear{manin2021running}) postulates that poetic devices expand the population due to the relaxation of language norms. A well-known metric that is used to test NLG output quality is perplexity, which is directly related to entropy. A study on the correspondence between perplexity and human evaluation revealed that creative output requires perplexity that is not too high and not too low \cite{keukeleire2020correspondence}. 

A recent study \cite{berns2020bridging} has introduced the concept of active divergence, which refers to the ability of a model to actively diverge from the training data to achieve a more original and diverse output, as opposed to the traditional method of training the model to perfectly mimic the training data. However, it may be more challenging for a language model to diverge from the training data while still adhering to the rules of grammar, semantics, and syntax. The sampling methods also play an essential role in the output predictability. Applying random sampling leads to a random non-sensical output, and greedy decoding tends to prefer banal or repetitive output. \citeauthor{holtzman2020curious} (\citeyear{holtzman2020curious}) noted that beam search decoding in NLG models may result in less surprising and more monotonic output compared to the high variability in human choice of words. In order to increase unpredictability while keeping randomness low, they proposed Nucleus Sampling, which randomly sample from the top-$k$ most probable words and set a probability threshold $p$ to ensure that sampling does not occur from peaked or flat distributions. Only the top-$p$ tokens are considered where $V^{(p)}$ is the smallest set of tokens such that: \[\sum\nolimits_{{x \in V^{(p)}}}^{} {P\left( {{x|x_{1:i-1}}} \right) \geq {{p}}}\]

The level of creativity in language models is influenced not only by the inference algorithms, but also by how similar the machine language modeling is to human language modeling. Deep learning models, particularly those that use transfer learning, have achieved impressive results on NLG tasks. Auto-regressive (AR) models, such as Generative Pre-Training models (GPT/2/3), generate text by sampling the next token from a probability distribution. Although humans may use causal language when they speak, they often try to modify their speech when they attempt to write creative text. Masked Language Models (MLM), such as BERT, predict masked tokens in a sentence based on context, which is similar to how humans try to guess a better word after removing it from a sentence and considering the context. However, these models will not automatically identify the words to remove and replace. Guided generation allows the model to be responsive and adaptable, and to accept or reject criteria, which may be a step towards a more social model. A study on the steerability of Generative Adversarial Networks (GANs) \cite{jahanian2019:steerability} shows that they can extrapolate yet are limited to the diversity of the data set. Previous NLG models have been limited by fixed or monotonically increasing sequence lengths, but Levenshtein Transformer \cite{gu2019} and Edit-based Transformer \cite{xu2021editor} allow for dynamic length changes. However, current large language models (LLMs) tend to overlook the integration of the unique elements associated with various forms of creative writing, like poetic diction, in their training process.

\section{Creative NLG: Devices and Tasks}
\label{section:4}
 
As discussed earlier, creativity in natural language appears in the original, rich and unusual use of language. Figurative language utilizes figures of speech to modify the literal meanings of words and produce imagery, or utilizes figures of sound to produce an appealing form. These rhetoric devices are used in creative writing in special ways to ensure novelty, richness and unpredictability. This includes similes, metaphors, hyperboles, sarcasm, humor, rhythms and rhymes, among others. These techniques are combined to shape the elements of various types of creative writing such as poetry and prose novels. While most research has focused on the detection of these literary devices, it was until recently that researchers showed interest in models that generate such devices. The employment of MLM and AR based models, along with the versatile BART model that amalgamates the strengths of both approaches and the use of COMET \cite{bosselut2019comet}, designed for generating commonsense knowledge, underpinned the observed trend. Next, we will review recent work and its impact on downstream NLG tasks before we introduce poetry generation.

\subsubsection{Metaphor:} Automated metaphor generation has only recently received increased attention \cite{tong2021recent}. MERMAID \cite{chakrabarty2021mermaid} fine-tunes BART with an automatically generated parallel corpus using an MLM model. The MLM will replace metaphorical verbs with their literal complement using the symbol relation from COMET. Then, they use a generator-discriminator approach to transform literal expressions to metaphorical ones with a verb-replacement objective and a top-$k$ sampling strategy. The generated sentences are scored using a RoBERTa-based metaphor detection model to favor sentences with higher quality metaphorical verbs. \citeauthor{stowe2021metaphor} (\citeyear{stowe2021metaphor}) propose a similar model that replaces literal verbs with metaphorical ones guided by conceptual mapping. The authors propose two methods, one by training a frame embedding model and the other by finetuning BART using a generated parallel corpus similar to the one used in MERMAID but tagging each sentence with FrameNet frame labels. A limitation of these models is that they only generate metaphorical verbs and assume a given context.

\subsubsection{Simile:} \citeauthor{chakrabarty2020generating} (\citeyear{chakrabarty2020generating}) finetune BART on an automatically built corpus that transforms similes to their literal compliments using the ``has property" relationship provided by COMET to generate similes from a given literal sentence by replacing a word by a novel simile. A limitation of the proposed model is that it explicitly replaces only an adjective or an adverb from a given input. \citeauthor{zhang2021writing} (\citeyear{zhang2021writing}) propose a simile insertion approach based on the general context of the sentence without replacing words. The approach consists of two main stages: (1) detecting an appropriate simile insertion position in the input sentence using a BERT-based model (2) generating a simile at that position using a standard transformer decoder. Both models are trained jointly in a multi-task learning setup.


\subsubsection{Hyperbole:} MOVER \cite{zhang2021mover} finetunes BART on a dataset of hyperboles retrieved from an online corpus using a BERT-based hyperbole detection model. The model is trained on a mask-filling task where potential hyperbolic tokens are identified, masked, and then regenerated based on their part-of-speech (POS) n-gram and ranked by an unexpectedness score. To generate a hyperbole, the model masks a non-hyperbole sentence span using a POS n-gram and then selects the highest-ranked of multiple possible hyperbole candidates using a BERT-based hyperbole ranker. HypoGen \cite{tian2021hypogen} focuses on clause-level hyperboles with a specific pattern of the form (A1 is so A2 that B is D). To train this model, the authors collected and annotated a dataset of hyperboles from Reddit posts (HYPO-Red), selecting a subset that contains the target pattern and analyzing the relationships between clauses. They then used COMET and reverse-COMET models to generate candidate clauses with commonsense and counterfactual relations, respectively, and used BERT-based and neural-based classifiers to select the top-$k$ candidates.

\subsubsection{Sarcasm:} To generate sarcasm, \citeauthor{chakrabarty2020r} (\citeyear{chakrabarty2020r}) negate or replace evaluative words with their antonyms and provide a context to emphasize the semantic incongruity between the intended sarcasm and the context. Several context candidates are retrieved using a concept derived from the COMET model and an online corpus. The candidates are ranked by a RoBERTa model for semantic incongruity and the highest scored are concatenated to the sarcastic sentence. \citeauthor{oprea2021chandler} (\citeyear{oprea2021chandler}) argue that it is not sufficient to only negate the literal meaning of a sentence to produce sarcasm and propose a framework that is based on the implicit display theory of sarcasm to generate a sarcastic response to an input. The framework identifies an ironic environment by negating the event in the input utterance. A COMET-based rule-based method is used to produce an insincere negative attitude response based on the input event and one of the following relations: the action needed to perform the event, the attribute needed to perform the action, the user reaction, and the effect of the action on the user. 

\subsubsection{Puns:} \citeauthor{yu2018neural} (\citeyear{yu2018neural}) use a conditional LSTM model to generate homographic puns given two senses of a pun word. They use a joint beam search decoding with a forward and backward generation centered by the target pun word. Both sequences are then concatenated to form the final pun sentence. AMBIPUN \cite{mittal2022ambipun} generates homographic pun sentences using GPT-3 to generate related context words given two different senses of a pun word. The context words are then combined using a T5 model to generate candidate sentences and a BERT-based model to rank and pick a final pun sentence. 

\subsubsection{Multi-Figurative:} Creative writing tasks need to employ various literary techniques in the generation. It is more practical to have one multi-figurative generation model to do so.
\citeauthor{lai2022multi} (\citeyear{lai2022multi}) propose a multi-figurative model that can transform between literal or figurative forms to another figurative form. The authors first finetune a BART model on a denoising objective to infuse multi-figurative sense into the model, then they fine-tune the model on a literal to figurative paraphrasing objective with a parallel data corpus and another time with a figurative to figurative paraphrasing objective with a cross attention layer to leak the target figure of speech to the encoder. For inference, they either transform the input text to the target figure of speech directly or to a literal form first and then to the target figurative form. 

\subsection{Impact on NLG Downstream Tasks}
Generating creative language plays an essential role in many text-to-text NLG tasks, especially in creative writing tasks such as poetry and prose novels or stories. It also improves the quality of downstream tasks such as machine translation, dialogue generation, text summarization, style transfer, and even other data-to-text applications.

\subsubsection{Storytelling} is to generate an open-ended text that conveys to the reader a comprehensive story. It is one of the creative writing tasks written in prose form as opposed to poetry. Storytelling needs to account for soft and hard constraints such as the topics, plots and characters among other aspects to keep the story consistent and coherent. AI-aided approaches were used \cite{goldfarb-tarrant2019a} to collaboratively write and plan story plots. Plug-and-Play Language Model (PPLM) \cite{dathathri2019plug} is used for controllable story generation by plugging an attribute classifier on top of a GPT-2 transformer that checks how similar the next token is to a given topic. The classifier will be exposed to the history of latent representations of the generated words, then repeatedly perform backward and forward passes through the classifier and the generator calculating gradients each time, updating the representations, and handing them back to the generator. This will steer the generation process towards the topics at inference time with minimal training. The work was limited to assistive stories generation, completing a story skeleton given a story theme or sentiment. \citeauthor{brahman2020modeling} (\citeyear{brahman2020modeling}) introduce multiple emotion-aware GPT-2-based models coupled with the COMET model to generate stories given a title and an emotional arc of the protagonist. Their work was the first to generate stories with an emotional trajectory plan. 

\subsubsection{Text Style Transfer} is gaining an increasing popularity in natural language generation \cite{garbacea2020neural}. Variational Auto-Encoders (VAEs) were used to rewrite modern text in Shakespeare style \cite{mueller2017sequence}. \citeauthor{riedl2020weird} (\citeyear{riedl2020weird}) used a transformer-based model (XLNet) to generate a parody of lyrics, changing the lyrics while preserving the rhythm and syllable count.

\subsubsection{Machine Translation} \citeauthor{low2011translating} (\citeyear{low2011translating}) examines the challenges and strategies involved in translating jokes and puns. The author argues that while some forms of humor are easily translatable, others, such as those that rely on wordplay and cultural references, can be particularly difficult. To effectively translate this kind of humor, the paper suggests using a variety of strategies, such as sense transferring, semantic leaps and cultural substitution. 

\subsubsection{Dialogue Generation} can be modulated to generate creative dialogue content such as improv games, battle rapping, interactional jokes and conversational narratives, among others. In fact, recent chatbots such as LaMDA \cite{thoppilan2022lamda} and ChatGPT\footnote{\url{openai.com/blog/chatgpt/}} have shown the ability to write and complete jokes, poems, and stories in conversations. 

\subsubsection{Text Summarization} is the task of condensing a longer text preserving the most important information within its content. Similarly, creative language generation can be seen as summarizing broad ideas and beautiful imagery using words. In fact, iPoet \cite{yan2013poet} formulates the poetry generation task as a text summarization task. Given users' intents, write a poem that obeys poem requirements by retrieving terms out of a poetry corpora.

\subsubsection{Data-to-text generation:} Creative data-to-text tasks are analogous to how humans describe, in creative writing, what they see or hear from natural beauty, scenery, natural soundscape, etc. 
\citeauthor{loller2018deep} (\citeyear{loller2018deep}) used Inception, ConceptNet and LSTM models to generate image-inspired poems. \citeauthor{liu2018beyond} (\citeyear{liu2018beyond}) proposed a multi-adversarial CNN-RNN-based GAN model to caption images with a poem. \citeauthor{uehara2022vinter} (\citeyear{uehara2022vinter}) proposed transformer-based encoder and decoder models to generate emotional narratives from visual embedding extracted from images using CNNs. \citeauthor{achlioptas2021artemis} (\citeyear{achlioptas2021artemis}) demonstrated the power of utilizing their explained emotion-captioned image data set in enhancing language models to express and explain emotions triggered by artistic images.  In addition, \citeauthor{2020chen} (\citeyear{2020chen}) proposed a SeqGAN-based model that generates a matching line of lyrics with an input melody. 

\section{Poetry Generation}
\label{section:5}

Modeling poetry is more difficult than language modeling as it requires the machine not only to understand but also to use language as a creative tool. 
As previously discussed, creativity is not yet well-defined, and it is uncertain if computers can achieve it. 

\subsubsection{Definition} 
Before discussing how to generate poetry, we first need to define poetry, which is equally difficult as defining creativity itself. In fact, there are as many ways to describe poetry as there are people \cite{murfin_ray_2009}. Poetry is often defined as a type of writing that uses distinctive style, rhythm, and language to convey intense feelings and ideas \footnote{ \url{oed.com/view/Entry/146552}} or the writing that concentrates imaginative experience to elicit a specific emotional response through meaning, rhythm, and sound patterns \footnote{\url{merriam-webster.com/dictionary/poetry}}. According to \citeauthor{milic1970possible} (\citeyear{milic1970possible}), poetry is the writing that violates the logical sequence and semantic categories of prose.
Both prose and poetry may use literary devices such as metaphors, similes, ironies, and puns to express ideas and evoke emotions. 
Poetry, however, employs specific, but not essential, devices such as rhyme, rhythm, and meter. Meanwhile, the most notable devices found in a poem are its versification and form coherence.

\subsubsection{Creativity Dimensions} Various definitions of poetry highlight the three dimensions of creativity previously mentioned, delivered through language. A poem is a language artifact that requires the use of language to express meaning, and thus must adhere to linguistic rules to some extent while also allowing for violations, commonly referred to as poetic license. Originality and unpredictability can be observed through the surprising deviation from prose and normal speech rules, as well as through the use of poetic devices and euphony. The social aspect of poetry can be emphasized through the concentration of ideas, imagery and emotions within the boundaries of verses and stanzas, as well as the ability for the poem to be subject to criticism. 

\subsubsection{Forms} English poetry can be classified based on various poetic characteristics.
Notably, rhythm and rhyme schemes are often used to classify poems. The most popular forms include: (1) \textit{Free Verse}: which has no constraints on a specific form. (2) \textit{Haiku}: a short form of poetry that follows a 5-7-5 syllabic pattern. (3) \textit{Sonnet}: characterized by specific rhyme and meter schemes. Shakespearean sonnet is considered one of the most well-known with 3 quatrains and a couplet: ABAB CDCD EFEF GG rhyme scheme. (4) \textit{Blank Verse}: consists of unrhymed but metered verses. (5) \textit{Limerick}: consists of five lines with seven to ten syllables with a verbal rhythm and the first, second, and fifth lines rhyming.

\subsubsection{Goals} The most common goals of writing poetry according to \citeauthor{preminger2016princeton} (\citeyear{preminger2016princeton}) are to imitate reality, attain special effects on the readers, communicate emotions, and be art for art's sake. Similarly, the automatic poetry generation must represent adequate computational goals. \citeauthor{milic1970possible} (\citeyear{milic1970possible}) upheld in an early view that the possible usefulness of computer poetry is to influence the doer (designer/end-user) to learn more about language, poets and poetry.

\begin{figure*}
	\centering
	\includegraphics[width=\textwidth]{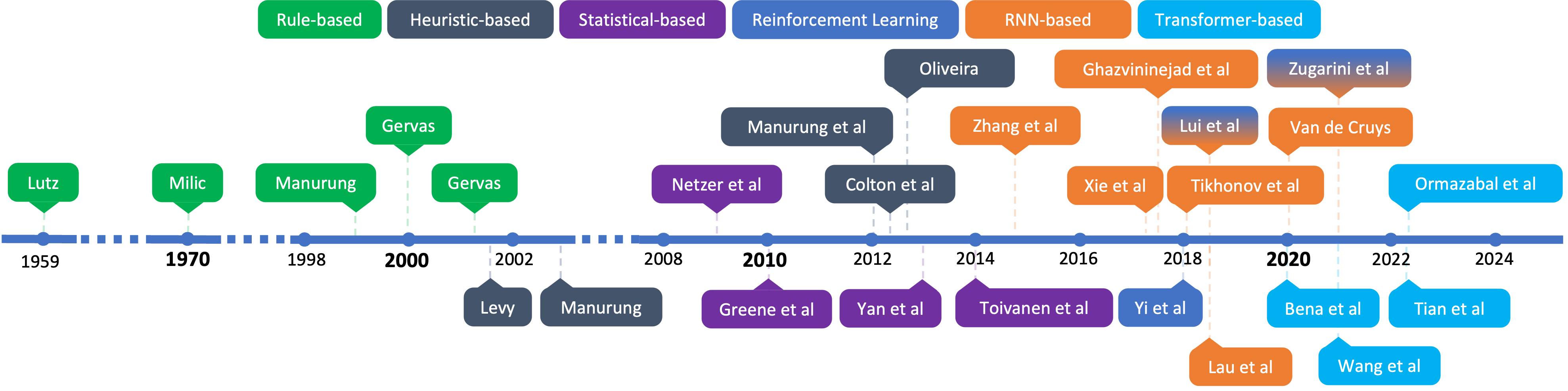}
        \caption{Poetry Generation Timeline.}
	\label{fig:timeline}
   	\vspace{-9pt}
\end{figure*}

\subsection{Poetry Generation Techniques}
Automatic poetry generation was documented as early as 1959 with a word salad approach by manually enhancing permutations of poem words \cite{lutz1959stochastische}. Since the few similar early attempts, the automatic poetry generation topic was inactive until the 1990s, where some works started getting attention in the field \cite{oliveira2009automatic}. Early poetry generation techniques can be categorized \cite{gervas2002exploring} into: template-based, generate-and-test, evolutionary, and case-based reasoning. \citeauthor{manurung2004evolutionary} (\citeyear{manurung2004evolutionary}) groups poetry generation approaches into: word salad, template/grammar-based, form-aware and poetry generation systems. \citeauthor{manurung2004evolutionary} argues that the first three methods do not satisfy the definition of poetry generators being: aware of meaningfulness, grammaticality and poeticness. \citeauthor{lamb2017taxonomy} (\citeyear{lamb2017taxonomy}) propose a generic categorization of poetry generation methods that works on any creative task: mere, human-enhanced, and computer-enhanced approaches. \citeauthor{oliveira2017survey} (\citeyear{oliveira2017survey}) reviews poetry generation methods from multiple intersecting angles: languages, forms, content quality, techniques, and evaluation. In addition, \citeauthor{oliveira2017survey} extends the list of techniques with: chart generation, constraint satisfaction, multi-agent, and language models such as Markov and RNN models. In this paper, we use a different taxonomy based on the intersection of chronology and methodology, pointing to the proposed creativity dimensions whenever applicable. We have found that the development of poetry generation techniques starts with rule-based, heuristic-based, statistical-based, then deep-learning-based approaches that are roughly proposed in this chronological order (Fig. \ref{fig:timeline}).

\subsubsection{Rule-based Approaches} 
The early methods are based on following crafted rules and filling poem templates. Templates are the poem shells where actual words are taken out from original poems, while some of their characteristics, such as part of speech, are preserved as placeholders to be systematically filled with other words matching the characteristics. RETURNER \cite{milic1970possible} is one of the earliest programs used to automatically generate English poems. The author designed three versions of a template-based model to generate three poems. The first version generates a poem given only the vocabulary of a real poem ``Return by Alberta Turner.'' The second version was based on subjects and verbs with randomly decided optional modifiers, complements, and end conjunctions to provide the opportunity for iteration. The third version is based on a six-slot grammatical matrix and a set of rules to guide the generation process. \citeauthor{manurung1999chart} (\citeyear{manurung1999chart}) uses a form-aware chart-based generation to generate strings that satisfy rhythm constraints. Meter validation is performed at each step of the process to reduce the search space. WASP \cite{gervas2000wasp} fills Spanish poem templates based on initial vocabulary and verse patterns, generating and validating verse drafts at each iteration according to length, rhyme, and rhythm constraints. The main focus of these early systems is on the poetic form rather than on the creative language. ASPERA \cite{gervas2001expert} is an evolved version of WASP with case-based reasoning to retrieve an existing form and vocabulary that are highly similar to the user-defined target specifications and then adapt to meet the desired specifications. However, this model is not fully automated and needs to manually interact with a human user to validate or correct the generated draft.
\subsubsection{Heuristic-based Approaches} 
Using rule-based approaches is advantageous as they provide control and adherence to specific poetic forms or templates, but they are very predictable. Heuristic methods, however, are less predictable, and unpredictability is essential to creative writing. \citeauthor{levy2001computational} (\citeyear{levy2001computational}) proposes a poet-critic model theory that consists of a computer-based generator and an evaluator module. The author did not implement the described theory but proposed a prototype that utilizes the concept of evolution to create original outputs and uses classical neural networks as evaluators trained on real-world human evaluations. The evolutionary algorithm is a meta-heuristic approach in which the best population of poems is chosen by evaluator modules, modified by crossover operators and mutation operators and continuously evaluated until a satisfactory result is obtained. \citeauthor{levy2001computational} uses such an approach to enhance poem features, words and rhymes. \citeauthor{manurung2004evolutionary} (\citeyear{manurung2004evolutionary}) uses a similar approach with the following operations: add, delete, or edit as forms of mutation. In a later update of the framework, \citeauthor{manurung2012using} (\citeyear{manurung2012using}) point to the fact that abusing the poetic license may lead to non-sense in a poetic form and that poetry generation approaches must conform to a restricted definition of poetry to generate falsifiable output. Similarly, \citeauthor{oliveira2012poetryme} (\citeyear{oliveira2012poetryme}) uses extracted semantic networks, grammar generators, and syllabic templates to generate Portuguese poems using genetic algorithms. \citeauthor{colton2012full} (\citeyear{colton2012full}) address the lack of story behind the generated poems. The authors avoid to retrieve from a corpus of existing poems to restrain plagiarism, instead they choose short similes and non-poetic text corpora. They combine extracted key phrases from the non-poetic corpus with produced variations of existing similes to fill a user-given template. They also attempt to automatically generate a commentary on the generation process by recording context statements at each step of the generation. 

\subsubsection{Statistical Approaches}
A problem in heuristic-based models is that they are not capable of modeling the language. \citeauthor{netzer2009gaiku} (\citeyear{netzer2009gaiku}) uses statistical methods to analyze a corpus of English Haikus and generate words related to a theme provided by a user. The authors pick a real Haiku poem POS template at random and fill-in words from the theme set. The sentences are then ranked according to the degree of association of words. In order to augment the unpredictability of the output, they give more weight to pick second degree association instead of first degree. \citeauthor{greene2010automatic} (\citeyear{greene2010automatic}) focus on the use of statistical methods to analyze, generate and translate poems. They use finite state transducers to map English sentences to a sequence of syllables and perform unsupervised learning to analyze the stress patterns of words in a poetry corpus. They used a finite state cascade to generate English love poems. In addition, the authors used a statistical phrase-based machine translation to translate Italian poems into English with the help of an iambic pentameter model. Poetry generation was also formulated as a text summarization task \cite{yan2013poet}: Given the intentions of a user, write a poem that meets the poem requirements by retrieving terms from a poetry corpora. The authors use generative summarization and aim to produce correlated verses taking into consideration the poetic form and semantic coherence. Word associations were also used to fill poetry templates \cite{toivanen2014officer} replacing words from a poem with other words based on syntactic similarity. They use the templates of poems from the Imagist movement to emphasize figurative language.

\subsubsection{Deep Learning Approaches}
Chomsky criticized statistical models as incomprehensible and inefficient compared to humans \cite{norvig2017}. The substantial amount of digital data created each second, the recent advances in computational power, and the ability of deep learning methods to derive complex data correlations, have led to state-of-the-art performance in NLG, including poetry generation. 

\subsubsection{• \textit{RNN-based}} RNNs were first introduced by John Hopfield in the early 1980s. However, they became popular as language models only a decade ago \cite{mikolov2010} for their ability to model long-term dependencies in sequential data such as text. \citeauthor{zhang2014chinese} (\citeyear{zhang2014chinese}) use a Convolutional Sentence Model (CSM) to capture sentence characteristics and use a character-based Recurrent Neural Network (RNN) model to incrementally generate poems. A set of keywords provided by the user is expanded with a poetic phrase taxonomy to generate candidate phrases, the top-ranked phrase will be the first line in the poem, and the following lines are sequentially generated given the previous line representations. The training process considers both the poetic form and semantics. \citeauthor{xie2017deep} (\citeyear{xie2017deep}) designed two RNN-based models where the first combines char-level and word-level LSTMs with GloVe word embeddings for the model to capture not only the semantics of the words, but also morphemes and rhymes during training. The second model uses CNNs to generate word embeddings. Both models showed significant performance in coherence, poeticness, and form characteristics over vanilla char-based and word-based RNN models. \citeauthor{ghazvininejad2017hafez} (\citeyear{ghazvininejad2017hafez}) proposed a sonnet generation system based on RNNs and finite state acceptors and used embedding rhymes with a word2vec model. The system is interactive and allows the user to input topic words, adjust style configurations, and evaluate the generated poem with a 5-star rating; the evaluation was used to update the system performance. \citeauthor{lau2018deep} (\citeyear{lau2018deep}) proposed a multi-model sonnet generator trained on Shakespeare’s sonnets by means of bidirectional word-based and char-based LSTMs with the attention mechanism. The model consists of language, rhythm and rhyme submodels with Bi-LSTMs with an attention mechanism; the paper showed that Bi-LSTMs can capture poetic forms such as rhythm and rhyme very efficiently. 
\citeauthor{tikhonov2018guess} (\citeyear{tikhonov2018guess}) used phoneme-based and char-based Bi-LSTMs, word-based LSTM, word embeddings and document embeddings for an author-stylized multilingual text generation framework with an application on English and Russian poems. 
\citeauthor{van2020automatic} (\citeyear{van2020automatic}) proposed an encoder-decoder model with double-layered GRUs using the attention mechanism. The decoder generated English and French poems line by line starting with the rhyming words backward to ensure a coherent verse with the forced rhyming scheme. To show the model's capability of presenting poeticness from scratch, the authors trained their model on non-poetic corpora. They proposed a latent semantic model to ensure topic coherence and a global optimization framework to score the verses and keep the highest-scoring ones.

\subsubsection{• \textit{RNNs and reinforcement learning (RL)}} RNNs are able to handle input sequences of variable length, but they may suffer from gradient vanishing during backpropagation, which can cause the loss of distant context in the generated text. Additionally, \citeauthor{yi2018automatic} (\citeyear{yi2018automatic}) argue that models based on maximum likelihood estimation, such as RNNs, tend to only optimize token-level loss rather than considering the poem as a whole, as humans do. To address this, they applied mutual RL to generate classic Chinese poems. The authors presented several reward modules instead of maximum likelihood to better mimic human behavior with reward schemes for fluency, coherence, meaning, and overall quality. \citeauthor{zugarini2021generate} (\citeyear{zugarini2021generate}) used deep RL by combining Bi-LSTM models and RL in an iterative refinement approach to generate and revise poems following an author style and a rhyme scheme. \citeauthor{liu2018beyond} (\citeyear{liu2018beyond}) also proposed a deep RL using a multi-adversarial CNN-RNN-based GAN model to caption images with a free verse English poem.
\subsubsection{• \textit{Transformer-based}} Transformers \cite{vaswani2017} were introduced to overcome RNNs limitations caused by their sequential design, by allowing parallelization of training. Instead of processing a sequence token by token, sequences are processed holistically with positional encoding. Additionally, transformers introduced self-attention, allowing inputs-to-inputs and outputs-to-outputs attention to enhance context encoding and decoding. \citeauthor{bena2020introducing} (\citeyear{bena2020introducing}) fine-tuned GPT-2 on a dream corpus to endow imagery language and fine-tuned it again on various emotion-based subcorpora to invoke one of the following emotions on the reader: joy, trust, anticipation, anger, fear, surprise, sadness, and disgust. LimGen \cite{wang2021there} used GPT-2 to predict the next words in multi-candidate templates extracted from human-written limerick, then filtered out the words for meter and rhyme consistency, and output the top N lines using a variation of beam search that calculated a diversity score between templates. \citeauthor{ormazabal2022poelm} (\citeyear{ormazabal2022poelm}) proposed a structure-aware training of an autoregressive transformer model to generate formal verse poems in Spanish and Basque with only non-poetic corpora.  Similarly, \citeauthor{tian2022zero} (\citeyear{tian2022zero}) proposed a multimodel framework trained on non-sonnet corpora to generate sonnets. The authors separated the training from the decoding and utilize a series of content planning, rhyme pair generation, polishing and constrained decoding to generate sonnets.  

\subsection{Recommendations}
A major concern in rule-based and heuristic-based methods is the originality of language output, particularly when using poem templates and simple synonym replacements. Similar concerns apply to models trained on poetry corpora and those using transfer learning to limit mimicking the language, style, theme, and semantics of the source. Further study on active divergence in NLG can potentially enhance originality, enabling the production of creative content that incorporates novel elements.
Another important area is optimizing decoding strategies through the use of enhanced sampling techniques, which can increase the unpredictability factor of the generated output. Recent advancements in improving nucleus sampling with randomized heads and introducing new sampling methods \cite{zhang2021lingxi} have shown promising results in terms of diversity and novelty of generated texts. To address the sociability aspect of poetry, we recommend the utilization of iterative refinement models guided by discriminator models or reinforcement learning models to provide feedback during the generation process in a poet-critic-like framework. Additionally, we recommend developing creative LLMs that integrate elements of poetic diction during pretraining and incorporating multi-figurative generation models to aid in effective poetry generation.

\section{Conclusion}
In this work, we bring attention to the problem of generating creative data for researchers interested in the field. We analyze natural language generation models based on the three dimensions of creativity we proposed: originality, sociability and unpredictability. We provide a brief overview of text generation tasks and their potential creative applications, as well as a review of recent works. Finally, we provide a comprehensive overview examination of the task of poetry generation, including the methodologies and models employed in the literature. We hope that this review encourages researchers and provides insights for those interested in creativity in natural language generation and poetry, particularly in considering the use of suggested creativity and poeticness criteria in language models.

\section{Acknowledgements}
We thank Michael Ullyot for his insightful feedback and the anonymous reviewers for their valuable input.

\bibliographystyle{iccc}
\bibliography{iccc}

\end{document}